\documentclass[journal]{IEEEtran}
\ifCLASSINFOpdf
\else
\fi
\usepackage{amsmath}
\usepackage{graphicx}
\usepackage{color}
\usepackage{balance,subfigure}
\usepackage{multirow}
\usepackage{rotating}
\usepackage{epstopdf}
\usepackage{wrapfig}
\usepackage{array}
\usepackage{cite}
\def\etal{et al.~}

\begin{document}
%
\title{End-to-end Audiovisual Speech Activity Detection with Bimodal Recurrent Neural Models}
%
%
%

\author{Fei Tao,~\IEEEmembership{Student Member,~IEEE,}
	Carlos~Busso,~\IEEEmembership{Senior Member, IEEE,}
	
	\thanks{S. Tao, and C. Busso are with the Erik Jonsson School of Engineering \& Computer Science, The University of Texas at Dallas, TX 75080  (e-mail:fxt120230@utdallas.edu, busso@utdallas.edu).}}   

%
%

\markboth{}%
{}
%



\maketitle

\begin{abstract}
\emph{Speech activity detection} (SAD) plays an important role in current speech processing systems, including \emph{automatic speech recognition} (ASR). 
SAD is particularly difficult in environments with acoustic noise. A practical solution is to incorporate visual information, increasing the robustness of the SAD approach. An audiovisual system has the advantage of being robust to different speech modes (e.g., whisper speech) or background noise. Recent advances in audiovisual speech processing using deep learning have opened opportunities to capture in a principled way the temporal relationships between acoustic and visual features. This study explores this idea proposing a \emph{bimodal recurrent neural network} (BRNN) framework for SAD. The approach models the temporal dynamic of the sequential audiovisual data, improving the accuracy and robustness of the proposed SAD system. Instead of estimating hand-crafted features, the study investigates an end-to-end training approach, where acoustic and visual features are directly learned from the raw data during training. The experimental evaluation considers a large audiovisual corpus with over 60.8 hours of recordings, collected from 105 speakers. The results demonstrate that the proposed framework leads to absolute improvements up to 1.2\% under practical scenarios over a VAD baseline using only audio implemented with \emph{deep neural network} (DNN). The proposed approach achieves 92.7\% F1-score when it is evaluated using the sensors from a portable tablet under noisy acoustic environment, which is only 1.0\% lower than the performance obtained under ideal conditions (e.g., clean speech obtained with a high definition camera and a close-talking microphone).
\end{abstract}


%
\IEEEpeerreviewmaketitle

\section{Introduction}
\label{sec:intro}

The success of voice assistant products including Siri, Google Assistant, and Cortana has made the use of speech technology more widespread in our life. These interfaces rely on several speech processing tasks, including \emph{speech activity detection} (SAD). SAD is a very important pre-processing step, especially for interfaces without push-to-talk button. The accuracy of a SAD system directly affects the performance of other speech processing technologies including \emph{automatic speech recognition} (ASR), speaker verification and identification, speech enhancement and speech emotion recognition \cite{Hinton_2012, Liu_2017, Parthasarathy_2017_3, Tao_2018_3}. A key challenge for SAD is the environmental noise observed in real world applications, which can greatly affect the performance of the speech interface, especially if the SAD models are built with energy-based features \cite{Pang_2017,Moattar_2009}.

An appealing way to increase the robustness of a SAD system against acoustic noise is to include visual features \cite{Potamianos_2017, Tao_2018_4}, mimicking the multimodal nature of speech perception used during daily human interactions \cite{Keil_2011, VanEngen_2017}. While this solution is not always practical for all applications, the use of portable devices with camera and the advances of \emph{human-robot interaction} (HRI) make an \emph{audiovisual speech activity detection} (AV-SAD) system a suitable solution. Noisy environment leads speakers to affect their articulatory production to increase their speech intelligibility, a phenomenon known as Lombard effect. While studies have reported differences in visual features between clean and noisy environments, these variations are not as pronounced as the differences in acoustic features \cite{Tran_2013}. Therefore, visual feature are more robust to acoustic noise. For example, Tao \etal \cite{Tao_2015} showed that a \emph{visual speech activity detection} (V-SAD) system can achieve robust performance under clean and noisy conditions using the camera of a portable device.

Conventional approaches to integrate acoustic and visual information in SAD tasks have relied on ad-hoc fusion schemes such as logic operation, feature concatenation or pre-defined rules \cite{Takeuchi_2009, Almajai_2008, Tao_2016, Petsatodis_2009}. These approaches oversimplify the relationship between audio and visual modalities, which may lead to rigid models that cannot capture the temporal dynamic between these modalities. Recent advances on \emph{deep neural network} (DNN) have provided new data-driven frameworks to appropriately model sequential data \cite{Hinton_2006, Bengio_2009}. These models avoid defining predefined rules or making unnecessary assumptions by directly learning relationships and distributions from the data \cite{Ngiam_2011}. Recent studies on audiovisual speech processing have demonstrated the potential of \emph{deep learning} (DL) in this area \cite{Tao_2018_4, Tao_2018_2, Petridis_2016, Petridis_2017, Chung_2017}. A straight forward extension from conventional approaches is concatenating audiovisual features as the input for a DNN \cite{Tao_2018_4,Ngiam_2011}. Another way is to rely on auto-encoder to extract bottleneck audiovisual representations \cite{Ariav_2018}. However, these methods do not directly capture the temporal relationship between acoustic and visual features. Furthermore, the systems still rely on hand-crafted features, which may not lead to optimal systems. 

This study proposes an end-to-end framework for AV-SAD that explicitly captures the temporal dynamic between acoustic and visual features. The approach builds upon the framework presented in our preliminary work \cite{Tao_2017,Tao_2018_5}, which relies on \emph{recurrent neural networks} (RNNs). Our approach, referred to as \emph{bimodal recurrent neural network} (BRNN), consists of three subsystems. The first two subsystems independently process the modalities using RNNs, creating an acoustic RNN and a visual RNN. These subsystems are implemented with \emph{long short-term memory} (LSTM) layers, and their objective is to capture the temporal relationship within each modality that are discriminative for speech activity. These subsystems provide  high level representations for the modalities, which are concatenated and fed as an input vector to a third subsystem. This system, also implemented with LSTMs, predicts the speech/non-speech label for each frame, capturing the temporal information across the modalities. 

An important contribution of this study is that the acoustic and visual features are directly learned from the data. Recent advances in DNN for speech processing tasks have shown the benefits of learning discriminative features as part of the training process, using \emph{convolutional neural network} (CNN) and sequence modeling with RNN \cite{Noda_2014,Graves_2006}. We can learn end-to-end system with this approach, which has led to performance improvements over hand-crafted features in many cases \cite{Petridis_2017,Zhang_2017_2}. Furthermore, we can capture the characteristics of the raw input data and extract discriminative representation for a target task \cite{Amodei_2016, Hannun_2014}. These observations motivate us to learn discriminate features from the data. The inputs of the BRNN framework are the raw image around the orofacial area as visual features, and the Mel-filterbank as acoustic features. For the visual input, we use three 2D convolutional layers to extract high-level representation from the raw image around the mouth area. On top of the convolutional layers, we use LSTM layers to model temporal information. For the acoustic input, we use \emph{fully connected} (FC) layers that are connected to LSTM layers to model the temporal evolution of the data, similar to the visual part. The proposed approach is jointly trained learning discriminative features from the data, creating an effective and robust end-to-end AV-SAD system.

We evaluate our framework on a subset of the CRSS-4English-14 corpus consisting of over 60h of recordings from 105 speakers. The corpus includes multiple sensors, which allows us to evaluate the proposed approach under ideal channels (i.e., close-taking microphone, high definition camera) or under more practical channels (i.e., camera and microphone from a portable tablet). The corpus also has noisy sessions where different types of noise were played during the recordings. The various conditions can mimic practical scenarios for speech-based interfaces. We replicate state-of-the-art supervised SAD approaches proposed in previous studies to demonstrate the superior performance of the proposed approach. The experimental evaluation shows that our end-to-end BRNN approach achieves the best performance under all conditions. The proposed approach can achieve at least 0.6\% absolute improvement compared to the state-of-the-art A-VAD system. Among the AV-SAD systems, the proposed approach outperforms the best baseline by 1.0\% in the most challenging scenario corresponding to sensors from a portable device under noisy environment. This result for this condition is 1.2\% higher than an A-SAD system, providing clear benefits of the proposed audiovisual solution for SAD.

The paper is organized as following. Section \ref{sec:relatedwork} reviews previous studies on AV-SAD, describing the differences with our approach, and highlighting our contributions. Section \ref{sec:data} describes the CRSS-4English-14 corpus and the post-processing steps to use the recordings. Section \ref{sec:framework} introduces our proposed end-to-end BRNN framework. Section \ref{sec:evaluation} presents the experimental evaluations that demonstrate the benefits of our approach. The paper concludes with Section \ref{sec:conclusion}, which summarizes our study and discusses potential future directions in this area.

\section{Related Work}
\label{sec:relatedwork}
A successful SAD system can have a direct impact on ASR performance by correctly identifying speech segments. While speech-based SAD systems have been extensively investigated, SAD systems based on visual features are still under development.

Visual information describing lip motion provides valuable cues to determine the presence or absence of speech. Several studies have relied on visual features to detect speech activity in speech-based interfaces \cite{Rivet_2007, DeCuetos_2000}. These methods either rely exclusively on visual features (V-SAD) \cite{Tao_2015,Navarathna_2011, Joosten_2013,Sodoyer_2006, Aubrey_2010,Aubrey_2007}, or complement acoustic features with visual cues \cite{Takeuchi_2009,Almajai_2008,Ahmad_2013}. As an emerging research topic, several methods have been proposed. Early studies relied on \emph{Gaussian mixture models} (GMM) \cite{Navarathna_2011,Liu_2004_2}, \emph{hidden Markov models} (HMMs) \cite{Aubrey_2007,Takeuchi_2009}, or static classifiers such as \emph{support vector machine} (SVM) \cite{Joosten_2013}. Recent studies have demonstrated the potential of using deep learning for V-SAD and AV-SAD \cite{Ariav_2018}.

The use of deep learning offers better alternatives to fuse audiovisual modalities. Early studies relied on simple fusion schemes, including concatenating audiovisual features \cite{Almajai_2008}, combining the individual SAD decisions using basic ``AND'' or ``OR'' operations \cite{Tao_2016,Takeuchi_2009}, and creating hierarchical decision rules to assess which systems to use \cite{Petsatodis_2009}. In contrast, DL solutions can incorporate in a  principled manner the two modalities. DL techniques can be used to build powerful data-driven frameworks, relying on the input data rather than rigid assumptions or rules \cite{Hinton_2006, Bengio_2009}. DL-based approaches provide better solutions for AV-SAD task compared with conventional approaches, increasing the flexibility of the systems.

The pioneer work of Ngiam \etal \cite{Ngiam_2011} demonstrated that DL can be a powerful tool for \emph{audiovisual automatic speech recognition} (AV-ASR). For AV-SAD, however, there are only few studies using DL approaches. One exception is the approach proposed by Ariav  \etal \cite{Ariav_2018}. They used an autoencoder to create an audiovisual bottleneck representation. The acoustic and visual features were concatenated and used as input of the autoencoder. The bottleneck features from the autoencoder were used as input of a RNN, which aimed to detect speech activity. This approach separated the fusion of audiovisual features (autoencoder) from the classifier (RNN), which may lead to a suboptimal system where the bottleneck features are not optimized for the SAD task. To globally optimize the fusion and temporal modeling, Tao and Busso \cite{Tao_2017, Tao_2018_5} proposed the \emph{bimodal recurrent neural network} (BRNN) framework for AV-SAD task. The approach used three RNNs as subnets following the structure in Figure \ref{fig:framework_brnn}. The framework can model the temporal information within and across modalities. The results show that this structure can outperform an RNN taking concatenated audiovisual features.

\subsection{Features for Speech Activity Detection}
\label{ssec:rw-features}

For acoustic features, studies have relied on \emph{Mel frequency cepstral coefficients} (MFCCs) \cite{Ryant_2013}, spectrum energy \cite{Pang_2017} and features describing speech periodicity \cite{Sadjadi_2013}. However, there is no standard set for visual features, where studies have proposed several hand-crafted features. For example, Navarathna  \etal \cite{Navarathna_2011} and Almajai and Milner \cite{Almajai_2008} used appearance-based features such as 2D \emph{discrete cosine transform} (DCT) coefficients from the orofacial area. Other studies have relied on geometric features \cite{Liu_2004_2,Petsatodis_2009,Aubrey_2007,Sodoyer_2006}. A common approach is to use \emph{active appearance model} (AAM) \cite{Aubrey_2007}. Tao and Busso \cite{Tao_2014} and Neti \etal \cite{Neti_2000} suggested that appearance based features have the disadvantage of being more speaker dependent, so using geometric features can provide representations with better generalization. Some studies have combined appearance and geometric features \cite{Tao_2014}.

Instead of using hand-crafted features, an appealing idea is to learn discriminative features from the data using end-to-end systems. The benefit of this approach is that the feature extraction and task modeling are jointly learned from the data, providing flexible and robust solutions. While this approach has been used in other areas, we are not aware of previous studies on end-to-end systems for AV-SAD. We hypothesize that this approach can lead to improvements in the performance, since feature representations obtained during the learning process have been shown to be effective on other tasks. CNN was originally used to learn features from images, since CNN can learn spatial, translation invariant representations from raw pixels \cite{LeCun_1998_2}. The spatial invariance property in CNN allows the system to learn robust high-level representations from the input data \cite{Krizhevsky_2012}. Saitoh \etal \cite{Saitoh_2016} used CNN to extract visual representation from concatenated images for \emph{visual automatic speech recognition } (V-ASR) task. Petridis \etal \cite{Petridis_2017} presented an end-to-end systems for V-ASR. They used raw images and their corresponding delta information as input to recognize words, relying on \emph{bidirectional LSTMs} (BLSTMs). Amodei \etal \cite{Amodei_2016} and Sercu \etal \cite{Sercu_2016} used CNN to extract high-level acoustic feature representations from raw acoustic data for \emph{audio automatic speech recognition} (A-ASR) tasks. In these studies, FC layers were stacked over the CNN, mapping the representation extracted by the CNNs into the classification task space (Soltau \etal \cite{Soltau_2014} showed the benefits of adding FC layers on top of CNN layers). Inspired by these studies, this study adopts a CNN as a feature extractor for visual features to learn high-level representations that are discriminative for AV-SAD tasks.

\subsection{Temporal Modeling for Speech Activity Detection}
\label{ssec:rw-temporal}

Speech is characterized by semi-periodic patterns that are distinctive from non-speech sounds such as laugh, lip-smack, and deep breath. The temporal cues are observed not only on speech features, but also on orofacial features reflecting the articulatory movements needed to produce speech. Therefore, modeling temporal information is very important for SAD. 

An approach to model temporal information is to include features that convey dynamic cues. A classic approach is by concatenating contiguous  frames, creating contextual feature vectors \cite{Ryant_2013,Navarathna_2011}. However, this approach relies on a pre-defined context window, which will constraint the capability of  static frameworks such as DNN. Temporal information can also be added by taking temporal derivatives of the features \cite{Liu_2004_2,Almajai_2008,Sodoyer_2006}, or by relying on optical flow features \cite{Aubrey_2010}. For example, Sodoyer \etal \cite{Sodoyer_2006} demonstrated that dynamic features extracted by taking derivatives are more effective than the actual values describing the lip configuration. Takeuchi \etal \cite{Takeuchi_2009} extracted the variance of optical flow as visual features to capture dynamic information. 

An alternative, but complementary, approach to model temporal information is by using frameworks that capture recurrent connections. A common approach in speech processing tasks is the use of RNNs, which rely on connections between two contiguous time steps capturing temporal dependencies in sequential signals \cite{Williams_1989,Mikolov_2010,Graves_2013, Bahdanau_2016}. Ariav \etal \cite{Ariav_2018} used RNNs to capture temporal information in a AV-SAD task. A popular RNN framework is the use of LSTM units, which have been successfully used for AV-SAD task, showing competitive performance \cite{Tao_2017, Tao_2018_5}. Our proposed approach build on the \emph{bimodal recurrent neural network} (BRNN) framework proposed in our previous studies \cite{Tao_2017, Tao_2018_5}, which is described in Section \ref{ssec:avrnn}.

\subsection{Contributions of this Study}
\label{ssec:contributions}

This study extends the BRNN framework proposed in Tao and Busso \cite{Tao_2017} by directly learning discriminative audiovisual features, creating an effective end-to-end AV-SAD system. While aspects of the BRNN framework were original presented in our preliminary work \cite{Tao_2017,Tao_2018_5}, this study provides key contributions. A key difference between this study and our previous work is the features used for the task. While our preliminary studies relied on hand-craft audiovisual features, our method learns discriminative features directly from the data. To the best of our knowledge, this is the first end-to-end AV-SAD system. 

Other key difference with our previous work is the experimental evaluation. The framework is exhaustively evaluated with other AV-SAD methods, obtaining state-of-the-art performance on a large audiovisual database. The study also demonstrates the benefits of using Mel-filterbank over speech spectrogram in the presence of acoustic noise. 

\section{The CRSS-4English-14 Audiovisual Corpus}
\label{sec:data}

\begin{figure}[tb]
\centering
\subfigure[Equipments]
{
 \includegraphics[width=.97\columnwidth]{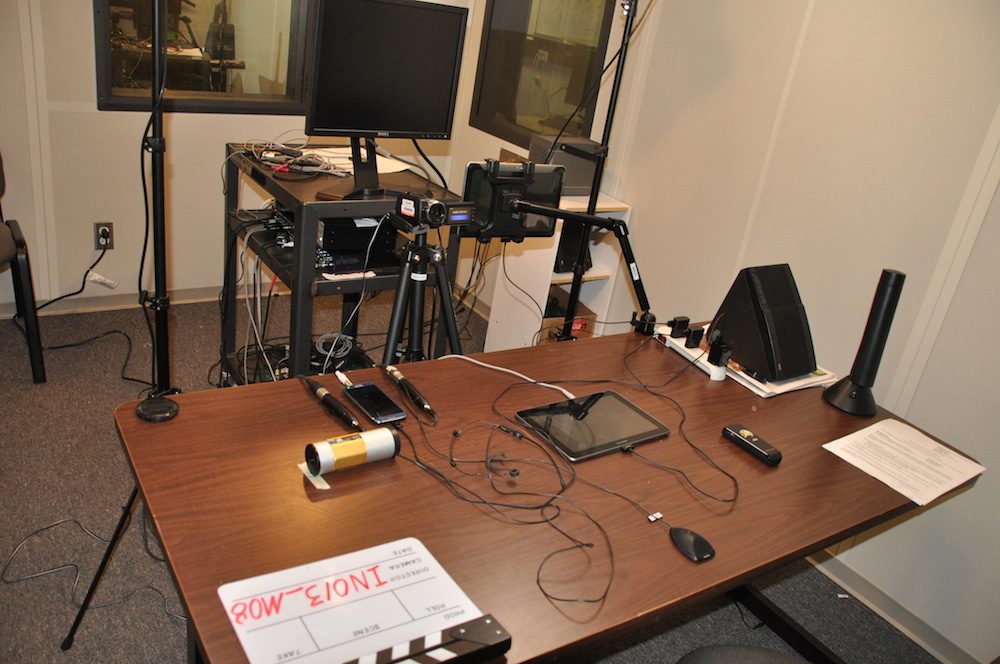}
\label{fig:crss_equipment}
}\hspace{0.01mm}
\subfigure[Recording setting]
{
 \includegraphics[width=.97\columnwidth]{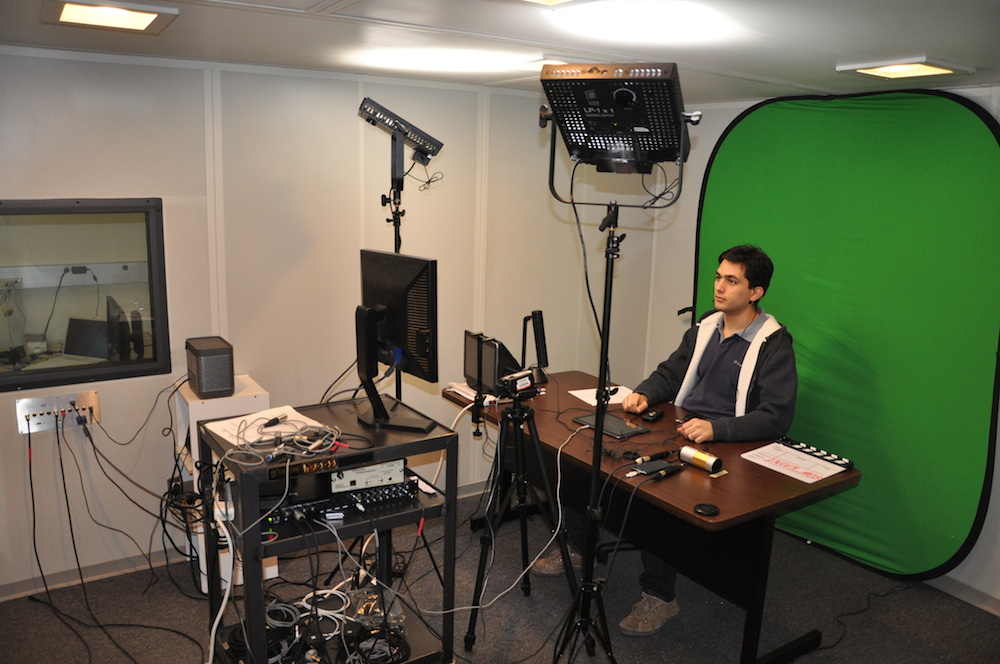}
\label{fig:crss_setting}
}\hspace{0.01mm}
\caption{Equipments and recording setting for the collection of the CRSS-4English-14 corpus. This study uses the audio from the close-taking and tablet microphones and the videos from the HD camera and the tablet.}
\label{fig:setting}
\end{figure}

This study uses the CRSS-4English-14 audiovisual corpus, which was collected by the \emph{center of robust speech systems} (CRSS) at the University of Texas at Dallas. The corpus was described in details in Tao and Busso \cite{Tao_2018_4}, so this section only describes the aspects of the corpus that are relevant to this study. Figure \ref{fig:setting} shows the settings used to collect the corpus. 

\subsection{Data Collection}
\label{ssec:datacollection}
The CRSS-4English-14 corpus was collected in a $13ft \times 13ft$ \emph{American Speech-Language-Hearing Association} ({ASHA}) certified sound booth. Figure \ref{fig:setting} shows the recording setting. The audio was collected at 48 kHz with five microphones: a close-talking microphone (Shure Beta 53), a desktop microphone (Shure MX391/S), the bottom and top microphones of a cellphone (Samsung Galaxy SIII), and a tablet (Samsung Galaxy Tab 10.1N). This study only uses the close-talking microphone, which was placed close to subject's mouth, and the microphone from the tablet, which was placed facing the subjects about two meters from them. The illumination was controlled with two LED light panels to collect high quality videos. The videos were collected with two cameras: a high definition (HD) camera (Sony HDR-XR100) and a camera from the tablet (Samsung Galaxy Tab 10.1N). This study uses the recordings from both cameras. The HD camera has a resolution of $1440 \times 1080$ at a frame rate of 29.97 fps. The tablet camera has a resolution of $1280 \times 720$ at a frame rate of 24 fps. Both cameras were placed facing the subjects about two meters from them, capturing frontal views of the head and shoulder of the subjects. A green screen was placed behind the subjects to create a smooth background. The participants were free to move their head and body during the data collection, without any constraint.

We used a computer screen about three meters from the subjects, presenting slides with the instructions for each task. The corpus includes read speech and spontaneous speech. For the read speech, we included different tasks such as single words (e.g., ``yes''), city names (e.g., ``Dallas, Texas''), short phrases or commands (e.g., ``change probe''), connected digit sequences (e.g., ``1,2,3,4''), questions (e.g. ``How tall is the Mountain Everest''), and continuous sentences (e.g., ``I'd like to see an action movie tonight, any recommendation?''), . For the spontaneous speech, we proposed questions to the speakers, who are required to respond using spontaneous speech. The sentences for each of the tasks are randomly selected from a pool of options created for the data collection, so the content per speaker is balanced, but not the same. The data collection starts with the \emph{clean session} where the speaker completed all the requested tasks (about 45 minutes). The clean session includes read and spontaneous speech. Then, we collected the \emph{noisy session} for five minutes. We played pre-recorded noise using an audio speaker (Beolit 12), which was located about three meters from the subjects. The noise recordings were recorded in four different locations: restaurant, house, office and shopping mall. Playing noise during the recording rather than adding artificial noise after the recording is a strength of this corpus, as it includes speech production variations associated with Lombard effect (speakers adapt their speech production in the presence of acoustic noise). For the noisy session, the slides were randomly selected from the ones used in the clean section. However, we only considered slides with read speech, discarding slides used for spontaneous speech.

The corpus contains recording from 442 subjects with four English Accent: American (115), Australian (103), Indian (112) and Hispanic (112). All the subjects are asked to speak in English. This study only uses the subset of the corpus collected from American speakers. During the recordings, we had problem with the data from 10 subjects, so we only use data collected from 105 subjects (55 females and 50 males). The total duration of this set is 60 hours and 48 minutes. The size, variability in tasks, presence of clean and noisy sessions, and use of multiple devices make this corpus unique to evaluate our AV-SAD framework under different conditions.

\subsection{Data Processing}
\label{ssec:dataprocessing}

A bell ring was recorded as a marker every time the subjects switched slides. This signal was used to segment the recordings. We manually transcribed the corpus, using forced-alignment to identify speech and non-speech labels. For this task, we use the open-source software SAILAlign \cite{Katsamanis_2011}. In the annotation and transcription of the speech, we annotate non-speech activities such as laughers, smacks, and coughs. We carefully remove these segments from frames labeled as `speech'.

We resample the sampling rate of the videos collected with the tablet to match the sampling rate of the HD camera (e.g., 29.97 fps). We also resample the audio to 16kHz for both audio channels (close-talking microphone and tablet).

\begin{figure}[t]
\centering
\includegraphics[width=0.8\columnwidth]{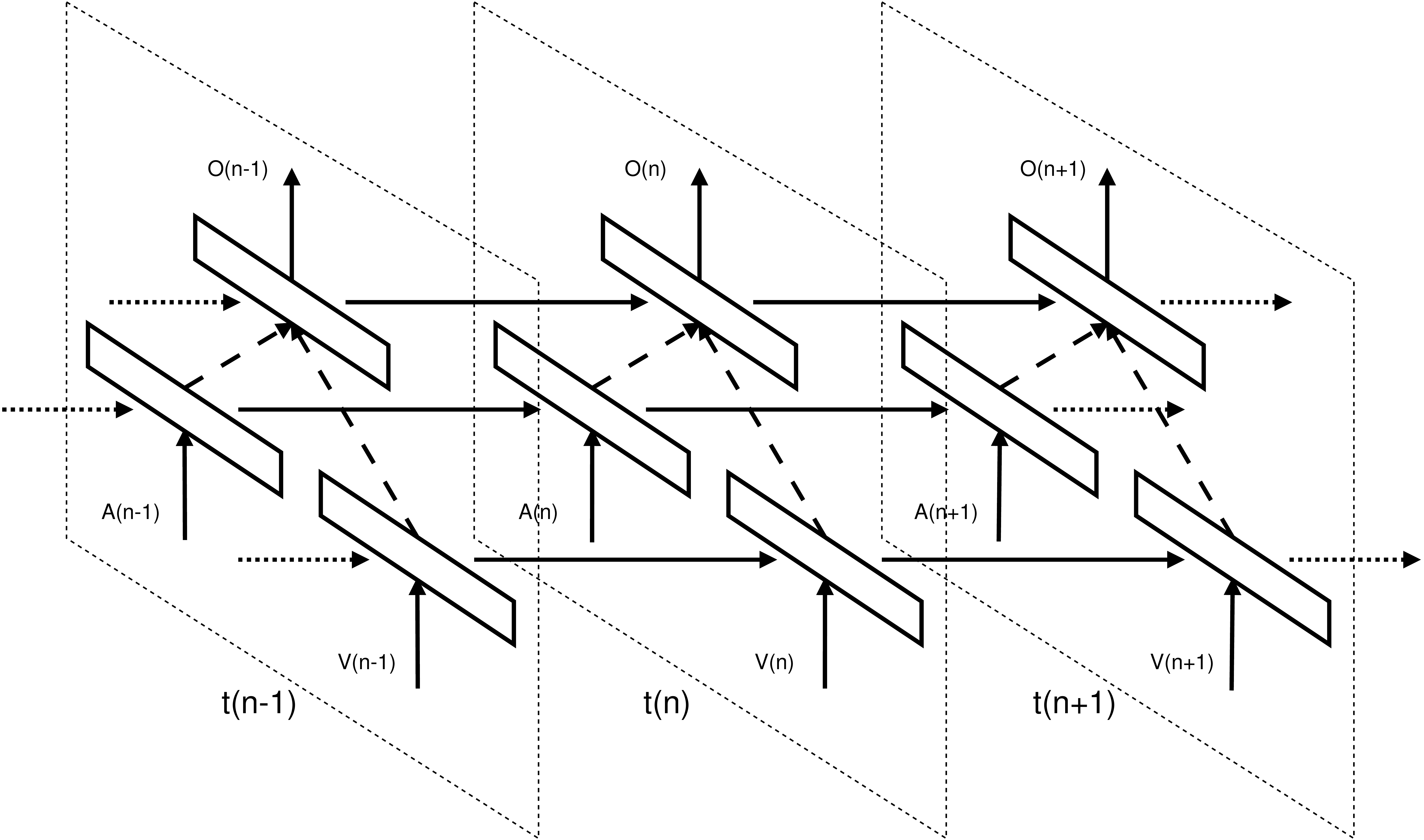}
\caption{Diagram of the BRNN framework, which consists of three subnets implemented with RNNs. The A-RNN processes acoustic information, while the V-RNN processes visual information. The AV-RNN takes the concatenation of the outputs from the A-RNN and V-RNN as input, predicting the task label as output.}
\label{fig:framework_brnn}
\end{figure}

\begin{figure}[t]
\centering
\subfigure[]
{
 \includegraphics[width=0.8\columnwidth]{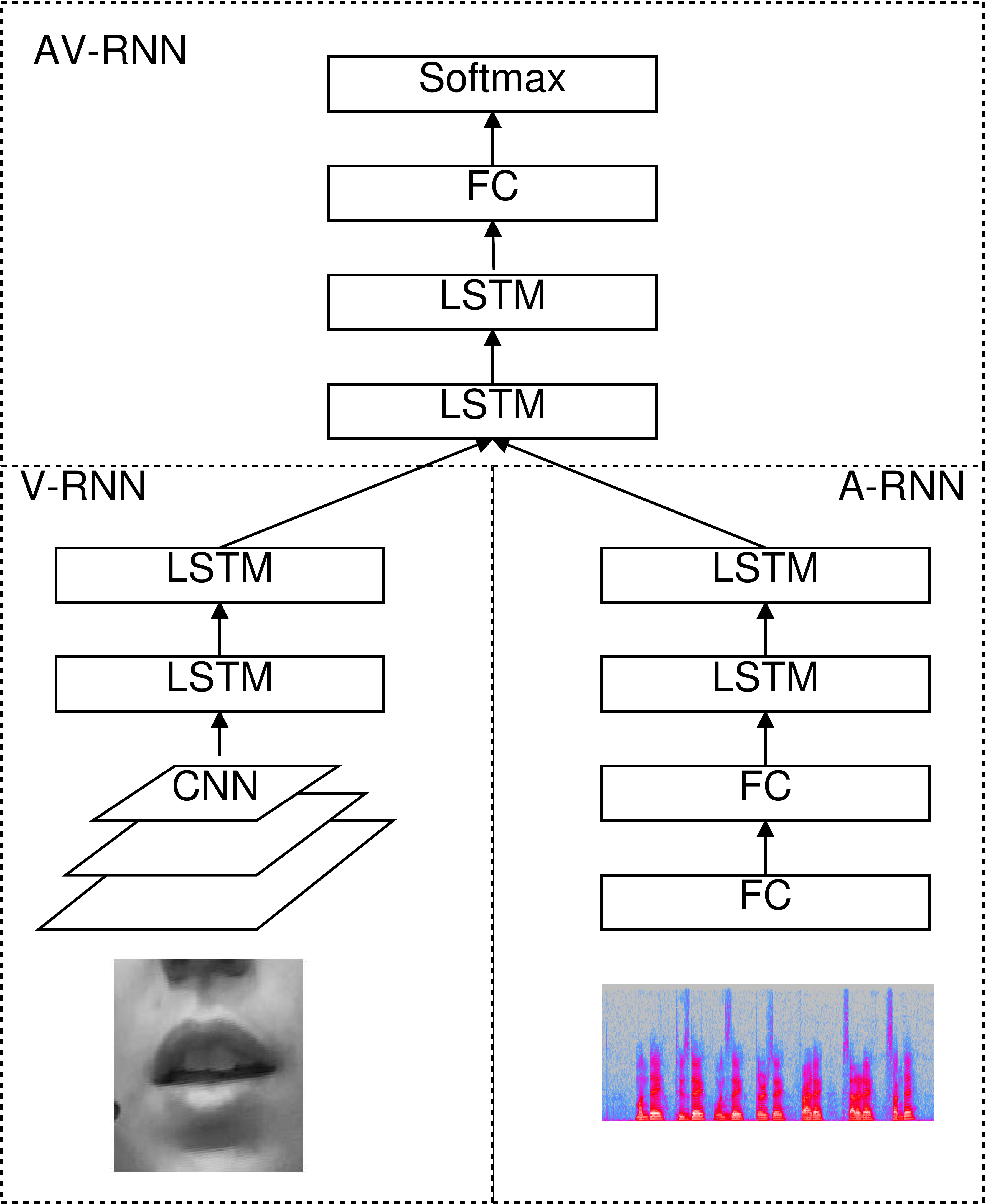}
\label{fig:framework_detail}
}\hspace{0.01mm}
\subfigure[]
{
 \includegraphics[width=1.0\columnwidth]{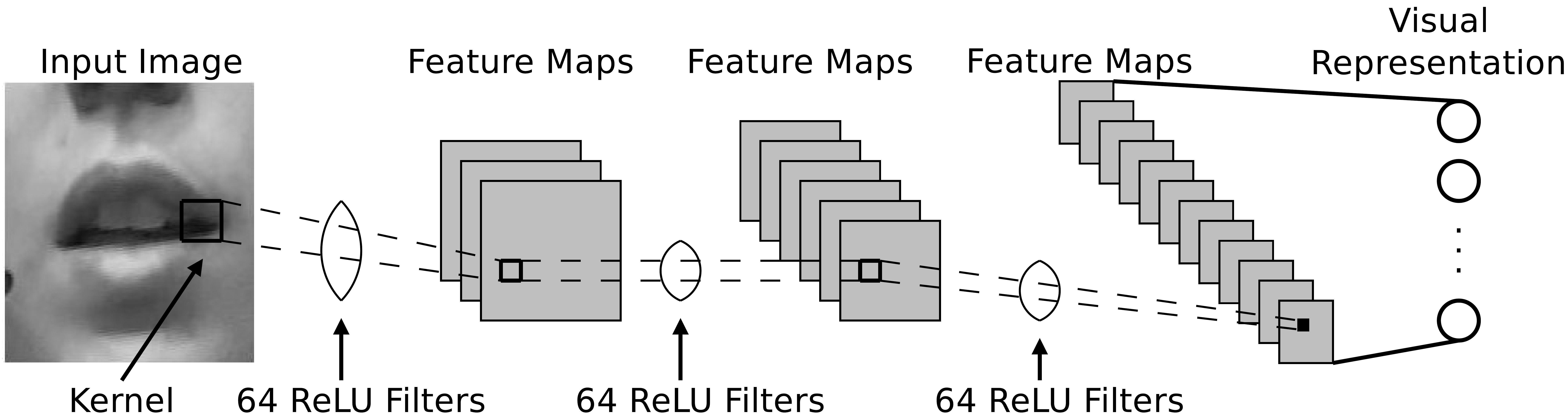}
\label{fig:framework_cnn}
}\hspace{0.01mm}
\caption{Detailed structure of the BRNN framework. (a) The structure of the proposed framework for one time frame. The A-RNN subnet includes FC and LSTM layers to process acoustic Mel-filterbank features. The V-RNN subnet has CNN extracting a visual representation from raw pixels and LSTMs to process temporal information. The AV-RNN subnet relies on FC and LSTM to process the concatenated output from the substructures A-RNN and V-RNN. (b) Detailed CNN configuration used to learn visual features.}
\label{fig:framework}
\end{figure}

\section{Proposed Framework}
\label{sec:framework}

In this study, we propose an end-to-end AV-SAD system building on the BRNN framework proposed in Fei and Busso \cite{Tao_2017}. Figure \ref{fig:framework_brnn} describes the BRNN framework, which has three subnets implemented with RNN: an audio subnet, a video subnet and an audiovisual subnets. The audio and video subnets separately process each set of the features, capturing the temporal dependencies within modality that are relevant for SAD. The outputs from these two RNNs are concatenated and fed into a third subnet, fusing the information by capturing the temporal dependencies across modalities. This section describes the three subnets. 

\subsection{Audio Recurrent Neural Network (A-RNN)}
\label{ssec:arnn}
The audio subnet corresponds to the \emph{audio recurrent neural network} (A-RNN) and it is described in Figure \ref{fig:framework_detail}. The A-RNN takes the acoustic features as input of a network consisting of static layers and dynamic layers (recurrent layers). The static layers model the input feature space, extracting the essential characteristics to determine speech activity. We rely on two \emph{fully connected} (FC) layers. This study uses two maxout layers \cite{Goodfellow_2014} rather than convolutional layers as static layers to reduce the computational complexity in training the models. Each layer has 512 neurons. The outputs of the FC layers are fed to dynamic layers to model the time dependencies within modality, as temporal patterns are important in SAD tasks \cite{Tao_2018_5}. We use two LSTM layers as dynamic layers. While bidirectional LSTMs have been used for this task \cite{Tao_2017}, we only use unidirectional LSTMs to reduce the latency of the model, as our goal is to implement this approach in practical applications. 

The acoustic feature used in our system are the Mel-filterbank features, which correspond to the energy in the frequency bands defined by the Mel scale filters. Therefore, it is a raw input feature that retains the main spectral information of the original speech. We use the tool \emph{python speech features}  to extract the mel-filterbank features, using the default setting (25 ms window, 10 ms shifting step, and 26D filters in the mel-filterbank). In this study, we concatenate 11 contiguous frames as input, which includes 10 previous frames, in addition to the current frame. Concatenating previous frames improves the temporal modeling of the framework, while keeping the latency of the system low.

\subsection{Visual Recurrent Neural Network (V-RNN)}
\label{ssec:vrnn}

The video subnet corresponds to the \emph{video recurrent neural network} (C-RNN) and it is also described in Figure \ref{fig:framework_detail}. The V-RNN takes raw images as visual feature. It extracts visual representation relying on convolutional layers. Figure \ref{fig:framework_cnn} describes the details of the architecture. The convolutional layers capture visual patterns, such as edges, corners and texture from raw pixels based on local convolutions. The visual patterns can depict the mouth appearance and shape associated with speech activity. We stack three convolutional layers with \emph{rectified linear units} (ReLUs) \cite{Nair_2010} to capture the visual patterns. Each layer has 64 filters. The kernel size is $5 \times 5$ and the stride is two (Fig. \ref{fig:framework_cnn}). By using stride, we reduce the size of the feature map, so we do not need to use the pooling operation. The feature representation defined by the CNNs is a 64D feature vector. The implementation is intended to keep a compact network with lower hardware requirements and computation workload, which is ideal for practical applications. On top of the convolutional layers, we rely on two LSTM layers to further process the extracted visual representation, capturing the temporal information along time. Each layer has 64 neurons. Therefore, the V-RNN is able to directly extract both the visual patterns and temporal information from raw images that are relevant for speech articulation.

\begin{figure}[t]
\centering
\includegraphics[width=\columnwidth]{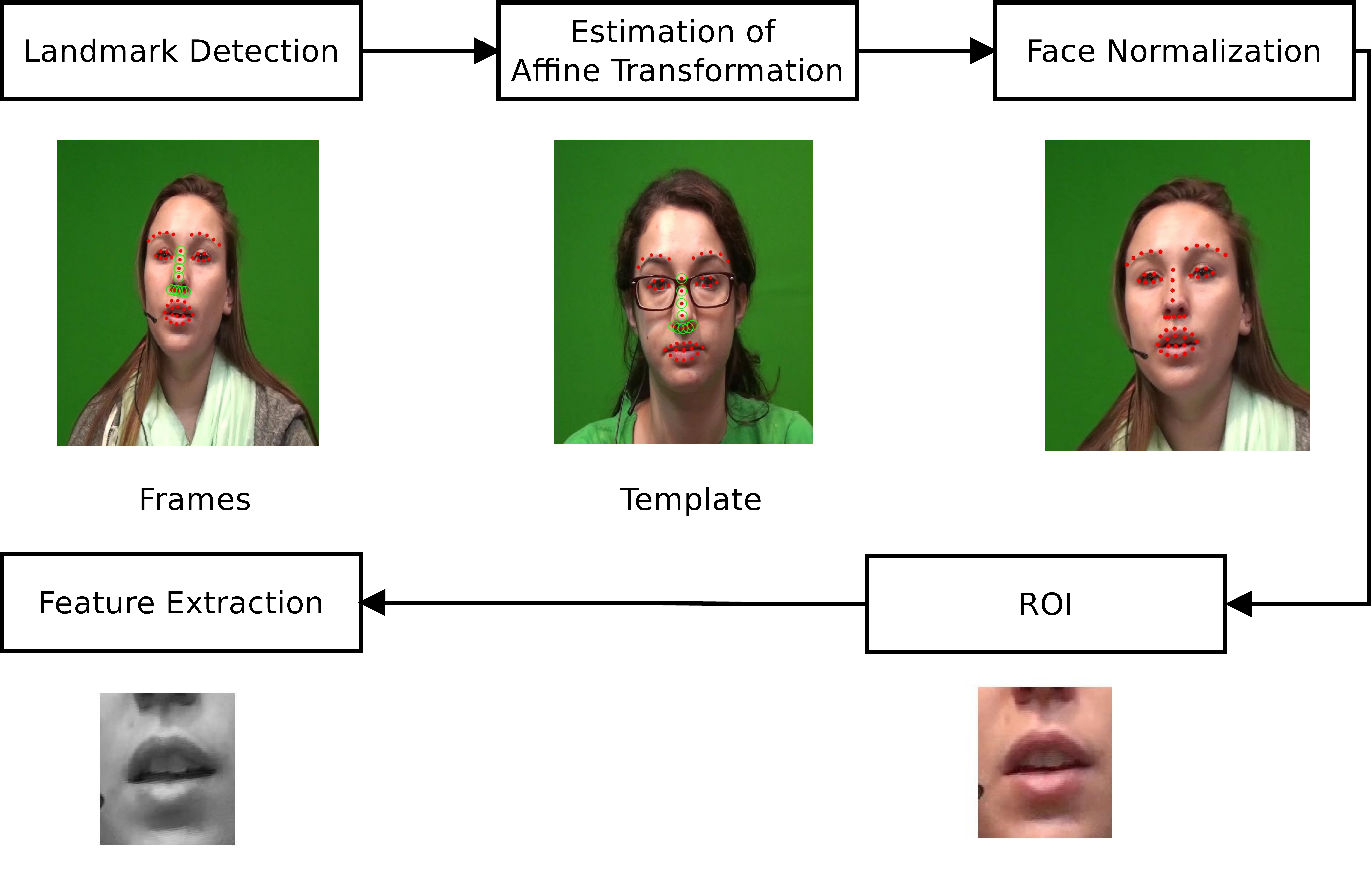}
\caption{Flowchart to extract the ROI around the orofacial area. The solid dots are facial landmarks. The dots with circle are used to estimate the affine transformation for normalization. The ROI is determined after normalization. The final mouth image is resized to $32\times32$ and transformed into a gray scale.}
\label{fig:extractfeat}
\end{figure}

Figure \ref{fig:extractfeat} shows the flow chart of the visual feature extraction process used in this study. We manually pick a frame of a subject posing a frontal face as the template. We extract 49 facial landmarks from the template and each frame of the videos. The facial landmarks are extracted with the toolkit IntraFace \cite{Xiong_2013}. IntraFace does not output a valid number when it fails to detect the landmarks, which occurred on some frames. We apply linear interpolation to predict these values when less than 10\% of the frames of a video are missing. Otherwise, we discard the video. We apply an affine transformation to normalize the face by comparing the positions of nine facial points in the template and the current frame. This normalization compensates for the rotation and size of the face. These nine points are selected around the nose area, because they are more stable when the people are speaking (points highlighted on Fig. \ref{fig:extractfeat} describing the template). After face normalization, we compute the mouth centroid based on the landmarks around mouth. We downsample the \emph{region of interest} (ROI) to $32\times32$ and convert it to gray scale colormap to limit the memory and computation workload.

\subsection{Audiovisual Recurrent Neural Network (AV-RNN)}
\label{ssec:avrnn}

The high-level feature representations provided by the top layers of the A-RNN and V-RNN subnets are concatenated together and fed into the \emph{audiovisual recurrent neural network} (AV-RNN) subnet (Fig. \ref{fig:framework_detail}). The proposed framework considers two LSTM layers to process the concatenated input. These LSTM layers aim to capture the temporal information across the modalities. On top of the LSTM layers, we include a FC layer implemented with maxout to further process the audiovisual representation. 
Each of the LSTM and FC layers are implemented with 512 neurons. The output is then sent to a softmax layer for classification, which determines whether the sample correspond to a speech or non-speech segment.

The BRNN framework is designed to model the time dependency within single modality and across the modalities. The convolutional layers allow us to directly extract visual representation from raw images. The framework also directly obtain acoustic representation from Mel-filterbank features. The proposed BRNN is jointly trained, minimizing a common objective function (in this case, the cross-entropy function). This framework provides a powerful end-to-end system for SAD, as demonstrated with the experimental results. 

\section{Experiments and Results}
\label{sec:evaluation}

\subsection{Experiment Settings}
\label{ssec:setting}

We evaluate our proposed approach on the CRSS-4English-14 corpus (Sec. \ref{sec:data}). We partition the corpus into train (data from 70 subjects), test (data from 25 subjects) and validation (data from 10 subjects) sets. All these sets are gender balanced. We use accuracy, recall rate, precision rate and F1-score as the performance metrics. The positive class to estimate precision and recall rates is \emph{speech} (i.e., frames with speech activity). We estimate F1-score with Equation \ref{eq:fscore}, combining precision and recall rates. 

\vspace{-0.2cm}
\begin{equation}
\mbox{F1-score} = 2 \times \frac{\mathit{Precision} \times \mathit{Recall}}{\mathit{Precision}+\mathit{Recall}}
\label{eq:fscore}
\end{equation}

We consider the F1-score as the main metric to compare alternative methods.We separately compute the results for each of the 25 subjects in the test set, reporting the average performance across individuals. We perform one-tailed t-test on the average of the  F1-scores to determine if one method is statistically better than the other, asserting significance at $p$-value$=$0.05. The experiments were conducted with GPU using the Nvidia GeForce GTX 1070 graphic card (8GB graphic memory).

Since all of the three baseline approaches rely on deep learning techniques, we apply the same training strategy (see sec \ref{ssec:baseline}). We use dropout with $p$= 0.1 to improve the generalization of the models. We use ADAM optimizer \cite{Kingma_2014_2}, monitoring the loss on the validation set. We rely on early stopping when the validation loss stops decreasing. 

\subsection{Baseline Methods}
\label{ssec:baseline}

We aim to compare the proposed approach with state-of-the-art methods proposed for SAD. The study considers three baselines. One approach only relies on acoustic features, and two approaches rely on audiovisual features. Our proposed approach uses unidirectional LSTM instead of BLSTM to reduce the latency in the model, which is a key feature for practical applications. To make the comparison fair, we also implement the baselines with unidirectional LSTMs. 


\subsubsection{A-SAD using DNN \cite{Ryant_2013}}
\label{sssec:Ryant}

The first baseline corresponds to the A-SAD framework proposed by Ryant \etal \cite{Ryant_2013}, which we denote ``Ryant-2013''. This approach is a state-of-the-art supervised A-SAD system using DNN. The system has four fully connected layers with 256 maxout neurons per layer. On top of the four layers, the system has a 2-class softmax layer for SAD classification. This approach uses 13D MFCCs. We concatenate 11 feature frames as input to make the system comparable with the proposed approach.

\subsubsection{AV-SAD system using BRNN \cite{Tao_2017}}
\label{sssec:Tao}

The second baseline is the AV-SAD system proposed by Tao and Busso \cite{Tao_2017}, which we denote ``Tao-2017''. This framework is a state-of-the-art AV-SAD system, relying on BRNN. The network is similar to the approach presented in this paper (Fig. \ref{fig:framework_brnn}). The key difference is the audiovisual features, which correspond to hand-crafted. 

The acoustic features correspond to the five features proposed by Sadjadi and Hansen \cite{Sadjadi_2013} for A-SAD: harmonicity, clarity, prediction gain, periodicity and perceptual spectral flux. These features capture key speech properties that are discriminative of speech segments such as periodicity and slow spectral fluctuations. \emph{Harmonicity}, also called \emph{Harmonics-to-Noise Ratio} (HNR), measures the relative value of the maximum autocorrelation peak, which produces high peaks for voiced segments. \emph{Clarity} is defined as the relative depth of the minimum \emph{average magnitude difference function} (AMDF) valley in the possible pitch range. This metric also leads to large values in the voiced segments. \emph{Prediction gain} corresponds to the energy ratio between the original signal and the \emph{linear prediction} (LP) residual signal. It will also show higher values for voiced segments. \emph{Periodicity} is a frequency domain feature based on the \emph{harmonic product spectrum} (HPS). \emph{Perceptual spectral flux} captures the quasi-stationary feature of the voice activity, as the spectral properties of speech do not change as quickly as non-speech segments or noise. The details of these features are explained in Sadjadi and Hansen \cite{Sadjadi_2013}. 

The visual features include geometric and optical flow features describing orofacial movements characteristic of speech articulation. We extract 26D visual features from the ROI shown in Figure \ref{fig:extractfeat}. This vector is created as follows. First, we extract a 7D feature vector from the ROI (three optical flow features, and four geometric features). The optical flow features consists of the variance of the optical flow in the vertical and horizontal direction within the ROI. The third optical flow feature corresponds to the summation of the variance in both direction, which provides the overall temporal dynamic on the frame. The four geometric features include the width, height, perimeter and area of the mouth. Based on the 7D feature vector, we compute three statistics over short-term window: variance, \emph{zero crossing rate} (ZCR) and \emph{speech periodic characteristic} (SPC) (details are introduced in Tao \etal \cite{Tao_2016}). The short-term window is shifted one frame at a time. We set its size equal to nine frames (about 0.3s) to balance the trade off between resolution (it requires short window) and robust estimation (it requires long window). The three statistics estimated over the 7D vector results in a 21D feature vector. We append the summation of the optical flow variances and the first order derivative of the 4D geometric feature to the 21D vector, since they can also provide dynamic information. The final visual feature is, therefore, a 26D vector. All the visual features are z-normalized at the utterance level.

We concatenate 11 audio feature frames as audio input, and use 1 visual feature frame as visual input. The subnet processing the audio features has four layers, each of them implemented with 256 neurons. The first two layers are maxout neuron layers and the other two layers are LSTM layers. The subnet processing the video features has four layers, each of them implemented with 64 neurons. The first two layers are maxout neuron layers and the last two layers are LSTM layers. The hidden values from the top layers of the two subnets are concatenated and fed to the third subnet, which has four layers. The first two layers are LSTM layers with 512 neurons. The third layer is implemented with maxout neurons with 512 neurons. The last layer is the softmax layer for classification. 

\subsubsection{AV-SAD System using Autoencoder \cite{Ariav_2018}}
\label{sssec:Autoencoder}

The third baseline correspond to the AV-SAD approach proposed by Ariav \etal \cite{Ariav_2018}, which relies on autoencoder (Section \ref{sec:relatedwork} describes this approach). We refer to this method as ``Ariav-2018''.
13D MFCCs are used as audio feature, and optical flow over the frame is used as visual feature. There are two stages in this approach. The first stage is the feature fusion stage which relies on an autoencoder. We implement this approach by concatenating 11 audio feature frames and one visual feature frame. The concatenated features are used as input for a five-layer autoencoder. The middle layer has 64 neurons, and the other layers have 256 neurons. The hidden values extracted from the middle layers are used as bottleneck features. All the layers use maxout neurons. The second stage uses the bottleneck features as the input of a four-layer RNN. The first two layers are LSTM layers, with 256 neurons per layer. The third layer is a maxout layer with 256 neurons. The last layer is the softmax layer for classification.

\subsection{Experimental Results}
\label{ssec:results}

While the CRSS-4English-14 corpus has several recording devices, this study only considers two combinations. The \emph{ideal channels} use the data collected with the close-talking microphone and the HD camera, which have the best quality. The \emph{practical channels} consider the video and audio recordings collected with the tablet. We expect that these sensors are good representations of the sensors used in practical speech-based interfaces. In addition, there are two types of environment as described in Section \ref{ssec:datacollection}: clean and noisy audio recordings. Altogether, this evaluation consider four testing conditions (ideal channel+clean; ideal channel+noise;  practical channel+clean; practical channel+noise). All the models are trained with the ideal channels under clear recordings, so the other conditions create train-test mismatches. We are interested in evaluating the robustness of the approaches under these channel and/or environment mismatches.

\begin{table}
\centering
\caption{Performance of the SAD systems for the ideal channels (close-talking microphone, HD camera). ``Env'' stands for testing environment (``C'' is clean; ``N'' is noisy). ``Modality'' stands for modality used by the approach (``A'' is A-SAD, ``AV'' is AV-SAD). ``Approach'' stands for corresponding framework (``Acc:'' accuracy; ``Pre:'' precision rate; ``Rec:'' recall rate;``F:'' F1-score).}
\begin{tabular*}{0.98\columnwidth}{@{\extracolsep{\fill}}c|c|c|c|c|c|c}
\hline
Env & Modality & Approach & Acc & Pre & Rec & F\\
\hline
\hline
\multirow{4}{*}{C} & A & Ryant-2013 & 90.3 & 96.6 & 90.5 & 93.4\\
& AV & Tao-2017 & 90.1 &  94.6 &  84.8 & 89.5\\
& AV & Ariav-2018 & 93.4 & 95.4 & 91.7 &  93.5\\
& AV & \textbf{Proposed} & 93.9 &  95.8 & 92.3 & 94.0\\ 
\hline
\multirow{4}{*}{N} & A & Ryant-2013 & 94.8 & 96.4 & 93.8 & 95.0\\ 
& AV & Tao-2017 & 93.3 & 93.1 & 94.0 & 93.4\\
& AV & Ariav-2018 & 94.4 & 95.4 & 94.1 & 94.7\\
& AV & \textbf{Proposed} & 95.3 & 96.2 & 95.2 & 95.7 \\
\hline
\end{tabular*}
\label{tab:performance_ideal}
\end{table}

The first four rows of Table \ref{tab:performance_ideal} shows the performance for the ideal channel under clean audio recordings. The proposed approach can outperform the baseline approaches (``Ryant-2013'' by 0.6\%; ``Tao-2017'' by 4.5\%; ``Ariav-2018'' by 0.5\%). The differences between our framework and ``Tao-2017'' are statistically significant ($p$-value $>$ 0.05). ``Tao-2017" used hand-crafted features.  Our end-to-end BRNN system achieves better performance, which demonstrate the benefits of learning the features from the raw data. The proposed approach perform slightly better than the baselines `Ryant-2013'' and ``Ariav-2018'', although the differences are not statistically significant.

\begin{figure}[t]
\centering
\subfigure[Ideal channels]
{
 \includegraphics[width=.97\columnwidth]{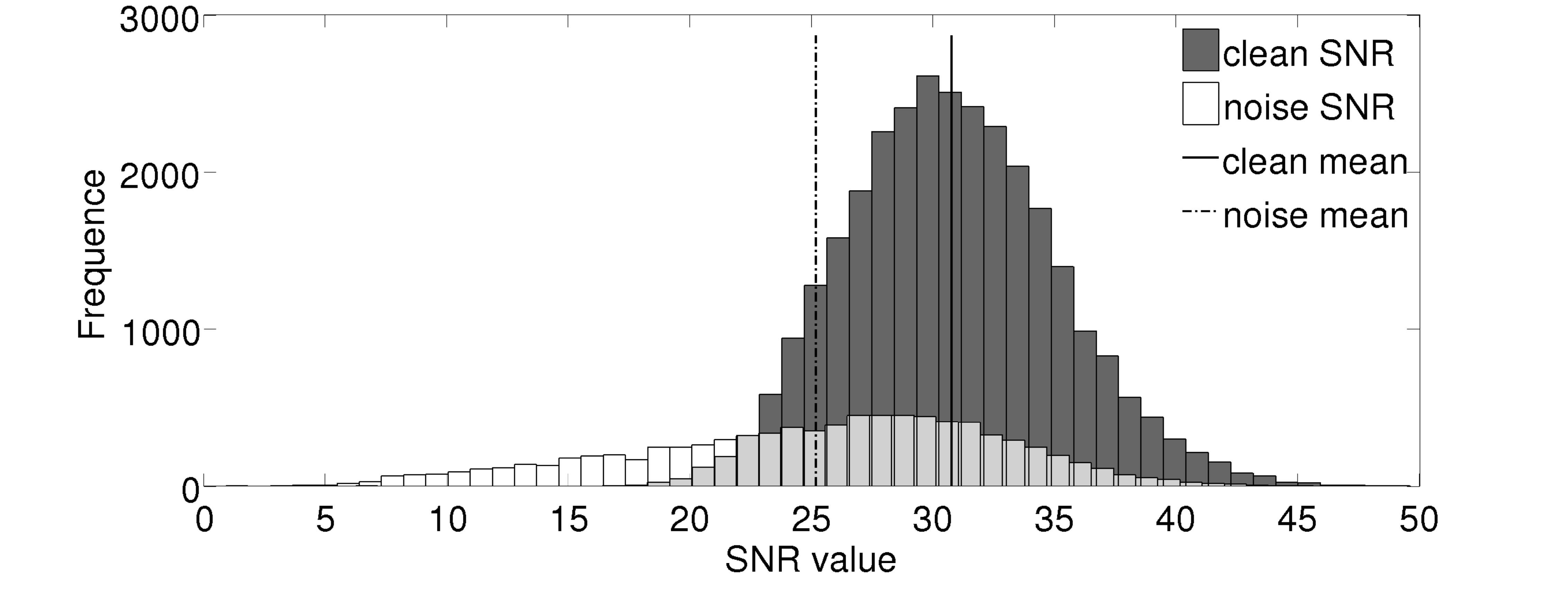}
 \label{fig:SNR_HD}
}\hspace{0.01mm}
\subfigure[Practical channels]
{
 \includegraphics[width=.97\columnwidth]{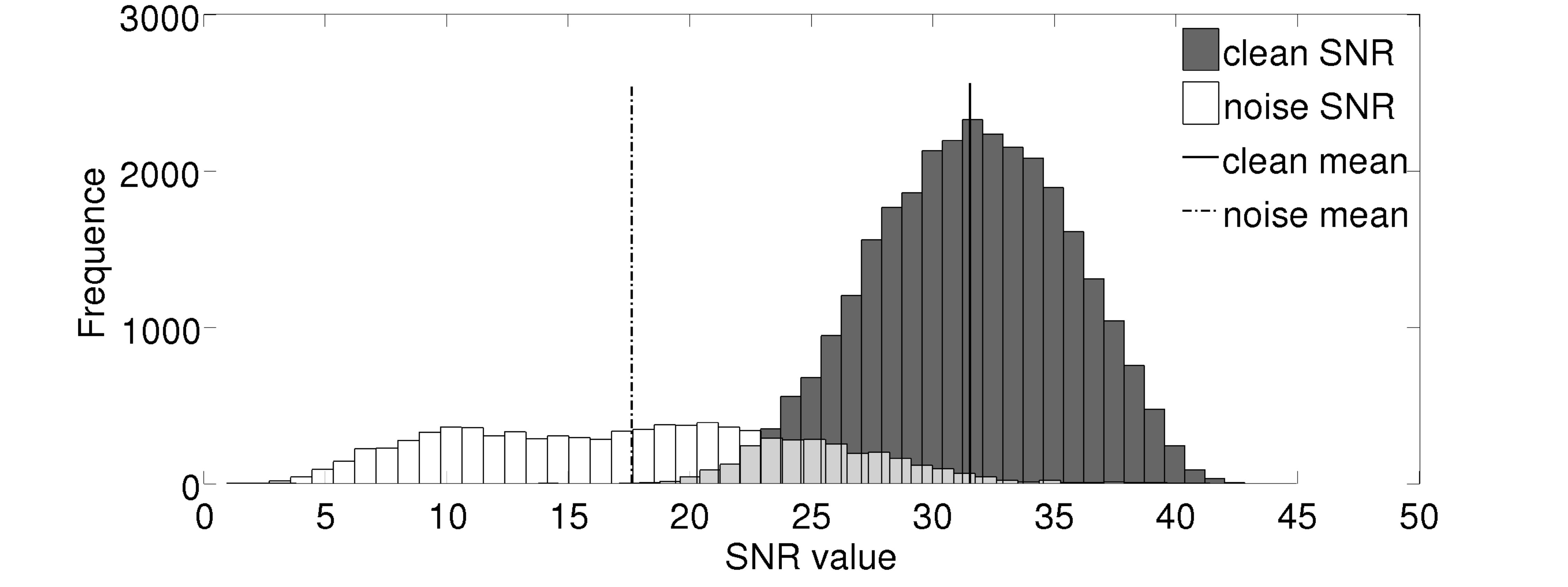}
 \label{fig:SNR_TG}
}\hspace{0.01mm}
\caption{Distributions of the SNR predictions for the ideal and practical channels. The SNR prediction are estimated with the NIST Speech SNR Toolkit \cite{Stanford_2005}(CRSS-4English-14 corpus). For the noisy audio recordings, the microphone in the tablet was closer to the audio speaker playing the noise, so the microphone of the practical channels is more affected by the noise.}
\label{fig:snr_new}
\end{figure}

The last four rows of Table \ref{tab:performance_ideal} shows the performance under 
noisy audio recordings. The proposed approach can outperform the baselines (``Ryant-2013'' by 0.7\%; ``Tao-2017'' by 2.3\%; ``Ariav-2018'' by 1.0\%). The differences are statistically significant ($p$-value$<$0.05) when our approach is compared with ``Tao-2017'' and ``Ariav-2018''. The classification improvement over the ``Ariav-2018" system demonstrates that the BRNN framework combine better the modalities than the autoencorder framework, especially in the presence of noise. The proposed BRNN structure jointly learns how to extract the features and fuse the modalities, improving the temporal modeling of the system. Table \ref{tab:performance_ideal} shows that the system tested with ideal channels has better performance under noisy conditions than under clean conditions. This unintuitive result is due to two reasons. First, the \emph{signal-to-noise ratios} (SNRs) under noisy and clean conditions are very similar for the ideal channels since the microphone is close to the subject's mouth and far from the audio speaker playing the noise. Figure \ref{fig:snr_new} shows  the distribution of the predicted SNR, using the NIST Speech SNR Toolkit \cite{Stanford_2005}. The Figure \ref{fig:SNR_HD} shows an important overlap between both conditions. Second, we only have read speech in the noisy section. In addition to read speech, the clean section also has spontaneous speech, which is a more difficult task for SAD.

\begin{table}
\centering
\caption{Performance of the SAD systems for the practical channels (microphone and camera from the tablet). ``Env'' stands for testing environment (``C'' is clean; ``N'' is noisy). ``Modality'' stands for modality used by the approach (``A'' is A-SAD, ``AV'' is AV-SAD). ``Approach'' stands for corresponding framework (``Acc:'' accuracy; ``Pre:'' precision rate; ``Rec:'' recall rate;``F:'' F1-score).}
\begin{tabular*}{0.98\columnwidth}{@{\extracolsep{\fill}}c|c|c|c|c|c|c}
\hline
Env & Modality & Approach & Acc & Pre & Rec & F\\
\hline
\hline
\multirow{4}{*}{C} & A & Ryant-2013 & 92.7 & 94.3 & 91.6 & 92.9\\ 
& AV & Tao-2017 & 90.0 &  91.9 & 87.3 & 89.4\\
& AV & Ariav-2018 & 92.8 & 95.2 & 90.8 & 92.9\\ 
& AV & \textbf{Proposed} & 93.4 & 95.4 & 92.0 & 93.7\\ 
\hline
\multirow{4}{*}{N} & A & Ryant-2013 & 90.8 & 90.6 & 92.5 & 91.5\\
& AV & Tao-2017 & 83.3 & 77.5 & 96.7 & 86.0\\
& AV & Ariav-2018 & 91.2 & 92.9 & 90.6 & 91.7 \\  
& AV & \textbf{Proposed} & 92.1 & 92.9 & 92.6 & 92.7\\ 
\hline
\end{tabular*}
\label{tab:performance_practical}
\end{table}

Table  \ref{tab:performance_practical} presents the results for the practical channels, which shows that our approach also achieves better performance than the baseline methods across conditions. For clean audio recordings, the proposed approach can significantly outperforms all the baselines (``Ryant-2013'' by 0.8\%; ``Tao-2017'' by  4.3\%; ``Ariav-2018'' by 0.8\%). For noisy audio recordings, we observe that the performances drop across conditions compared to the results obtained under clean recordings. The microphone of the tablet is closer to the audio speaker playing the noise, so the SNR is lower (Fig. \ref{fig:SNR_TG}). The proposed approach can maintain a 92.7\% F1-score performance, outperforming all the baseline frameworks (``Ryant-2013'' by 1.2\%; ``Tao-2017'' by  6.7\%; ``Ariav-2018'' by 1.0\%). The differences are statistically significant for all the baselines. This result shows that the proposed end-to-end BRNN framework can extract audiovisual feature representations that are robust against noisy audio recordings. 


\subsection{BRNN Implemented with Different Acoustic Features}
\label{ssec:AcousticFeatures}
We also re-implement the proposed approach with alternatives acoustic features to demonstrate the benefits of using Mel-filterbank features. The first acoustic features considered in this section is the spectrogram features without using the Mel filters. We extract 320D features using a Turkey filter with uniform bins between 0-8KHz. The second acoustic features correspond to the 5D hand-crafted acoustic features proposed by Sadjadi and Hansen \cite{Sadjadi_2013}, which we describe in Section \ref{sssec:Tao}. In both cases, we concatenate 10 previous frames to the current frame to create a contextual window, following the approach used for the Mel-filterbank features. For the spectrogram, the A-RNN subnet has four layers, each of them implemented with 4,096 neurons. For the 5D hand-craft features, the A-RNN subnet has four layers, each of them implemented with 256 neurons. The configuration for the rest of the framework is consistent with the proposed approach, including the A-RNN and AV-RNN. The evaluation only considers two conditions: ideal channels with clean audio recordings, and practical channels with noisy audio recordings. These two conditions represent the easiest and hardest settings considered in this study, respectively.

\begin{table}
\centering
\fontsize{9}{11}\selectfont
\caption{Performance of the BRNN framework implemented with different acoustic features. ``CH'' stands for channel. ``Env'' stands for testing environment (``C'' is clean; ``N'' is noisy). ``Feature'' stands for acoustic feature used in the evaluation (``Acc:'' accuracy; ``Pre:'' precision rate; ``Rec:'' recall rate;``F:'' F1-score).}
\begin{tabular*}{1.00\columnwidth}{@{\extracolsep{\fill}}c|c|c|c|c|c|c}
\hline
CH & Env & Feature & Acc & Pre & Rec & F\\
\hline
\hline
\multirow{4}{*}{Ideal} & \multirow{4}{*}{C} & Mel-filterbank & 93.8 &  95.8 & 92.3 & 94.0\\
& & Spectrogram & 93.4 & 94.8 & 93.1 & 93.9\\
& & Hand-crafted \cite{Sadjadi_2013}  & 92.2 & 94.0 & 90.4 & 92.2\\ 
\hline
\multirow{4}{*}{Practical} & \multirow{4}{*}{N} & Mel-filterbank & 92.1 & 92.9 & 92.6 & 92.7\\
& & Spectrogram & 76.8 & 71.3 & 97.4 & 82.2\\
& & Hand-crafted \cite{Sadjadi_2013} & 66.9 & 64.3 & 88.1 & 74.3\\ 
\hline
\end{tabular*}
\label{tab:acoufeat_comp}
\end{table}

Table \ref{tab:acoufeat_comp} presents the results. For the ideal channels under clean audio recordings, using a feature representation learnt from Mel-filterbank is slightly better than using a representation learnt from the spectrogram. Both of these feature representations lead to significantly better performance than the system trained with hand-crafted features. Learning flexible feature representations from the raw data lead to better performance than using hand-crafted features, as they are not constrained by pre-defined rules or assumptions. For the practical channels under noisy audio recordings, the model trained with hand-crafted features achieve the worse performance. The feature representation learnt from the Mel-filterbank is able to significantly outperforms the representation learnt from the spectrogram by a large margin (10.5\%). The feature representation learnt from the spectrogram is more sensitive to acoustic noise. 

The experiments in this section show that learning feature representations from Mel-filterbank leads to better performance across conditions, showing competitive results under clear and noisy speech.

\subsection{Performance of Unimodal Systems}
\label{sec:unimodal}
We also explore the performance of SAD systems trained with unimodal features to highlight the benefits of using audiovisual information. The experimental setup uses the V-RNN and A-RNN modules of the BRNN framework (Fig. \ref{fig:framework_detail}). For the audio-based system, we use the pre-trained A-RNN models. The weights of this subnet are not modified. On top of the A-RNN, we implement the same structure used in the BRNN consisting of two LSTM layers, one FC layer, and a softmax layer (Fig. \ref{fig:framework_detail}). We train these four layers from scratch, using the same training scheme used to train the BRNN network (i.e., dropout, ADAM, early stopping). The visual-based system is trained using the same strategy, starting with the V-RNN subnet. Similar to Section \ref{ssec:AcousticFeatures}, the evaluation only considers the ideal channels with clean audio recordings, and practical channels with noisy audio recordings.

\begin{table}
\centering
\caption{Performance for unimodal SAD systems and the bimodal SAD system. `CH'' stands for channel. ``Env'' stands for testing environment (``C'' is clean; ``N'' is noisy). ``Modality'' stands for modality used by the approach (``Acc:'' accuracy; ``Pre:'' precision rate; ``Rec:'' recall rate;``F:'' F1-score).}
\begin{tabular*}{0.98\columnwidth}{@{\extracolsep{\fill}}c|c|c|c|c|c|c}
\hline
CH & Env & Modality & Acc & Pre & Rec & F\\
\hline
\hline
\multirow{4}{*}{Ideal} & \multirow{4}{*}{C} & Bimodal & 93.8 &  95.8 & 92.3 & 94.0\\
& & Audio & 92.7 & 94.5 & 91.4 & 92.8\\
& & Video & 60.0 & 65.3 & 50.9 & 57.2\\ 
\hline
\multirow{4}{*}{Practical} & \multirow{4}{*}{N} & Bimodal & 92.1 & 92.9 & 92.6 & 92.7\\
& & Audio & 90.3 & 89.2 & 93.9 & 91.5\\
& & Video & 65.5 & 69.3 & 68.5 & 68.9\\ 
\hline
\end{tabular*}
\label{tab:uni}
\end{table}

Table \ref{tab:uni} shows the performance for the audio-based and visual-based systems. For comparison, we also include the proposed BRNN approach. For the ideal channels under clean audio recordings, the bimodal system can outperform the unimodal systems, showing the benefits of using audiovisual features. The result from the audio-based system is 35.6\% (absolute) better than the result from the video-based systems. This result is consistent with findings from previous study \cite{Tao_2017,Tao_2016,Tao_2015}. In spite of the lower performance of the visual-based system, the addition of orofacial features lead to clear improvements in the BRNN system. For noisy channels under noisy audio recordings, the bimodal system still achieves the best performance, outperforming the unimodal systems where the differences are statistically significant. The audio-based system achieves better results than the visual-based system. The performance for the visual-based system using noisy  audio recordings is higher than the results obtained with clean audio recordings. This result is  explained due to two reasons: (1) the visual features are not greatly affected by the background acoustic noise, and (2) the data for noisy audio recordings does not contain spontaneous speech, as explained in Section \ref{ssec:results}. If we include only read sentences recorded in both noisy and clear recordings, the performance of the visual-based system trained with the ideal channels under clean audio recordings is 69.2\%. This result is slightly higher than the value reported in Table \ref{tab:uni} for visual-based system under noisy audio recordings. 

The comparison between bimodal and unimodal inputs shows the benefit of using bimodal features. It highlights that our proposed BRNN approach can achieve better performance than state-of-the-art unimodal SAD systems.

\section{Conclusion and Future Work}
\label{sec:conclusion}

This study proposed an end-to-end AV-SAD framework where the acoustic and visual features are directly learnt during the training process. The proposed approach relies on LSTM layers to capture temporal dependencies within and across modalities. This objective is achieved with three subnets. The first two subnets separately learn visual and acoustic representations that are discriminative for SAD tasks. The visual subnet uses CNNs to learn features directly from images of the orofacial area. The audio subnet extracts acoustic representation directly from Mel-filterbank features.  The outputs of these subnets are concatenated and used as input of a third subnet, which models the temporal dependencies across modalities. Instead of using BLSTM, the proposed framework relies on unidirectional LSTM, reducing the latency, and, therefore, increasing the usability of the system in real applications. To the best of our knowledge, this is the first end-to-end AV-SAD system.

We evaluated the proposed approach on a set of the CRSS-4English-14 corpus (105 speakers), which is a large audiovisual corpus. The proposed approach outperformed alternative state-of-the-art A-SAD and AV-SAD systems. We observed consistent improvements across conditions. The proposed end-to-end BRNN framework maintained good performance in the presence of different noise and channel conditions. The system also achieved better performance than an implementation of the BRNN system using audiovisual hand-crafted features. These results demonstrated the benefits of learning feature representations during the training process. This approach provides an appealing solutions for practical applications. 

There are several research directions to extend the proposed approach. This study only focused on acoustic noise. In the future, we will evaluate the framework in the presence of visual artifacts (e.g., blurred images, occlusions). Likewise, the proposed approach assumes that the audiovisual modalities are available. We are exploring alternative solutions to address missing information. Finally, we leave as a future work to learn acoustic representations with CNNs. This direction was not pursued on this study due to the high computational cost required to train the models.

\section*{Acknowledgment}
This study was funded by the National Science Foundation (NSF) grants IIS-1718944 and  IIS-1453781 (CAREER).

\ifCLASSOPTIONcaptionsoff
  \newpage
\fi



%
\bibliographystyle{IEEEtran}
\bibliography{reference}

%

\vspace{-1.2cm}




\end{document}